%% file: fullpaper.tex
\documentclass{midl}

\usepackage{tikz, pgfplots}
\jmlrvolume{}
\jmlryear{}
\jmlrworkshop{}
\editors{}
\usepackage{siunitx}
\usepackage[version=3]{mhchem}

\title[Diffusion-Based Segmentation]{Diffusion Models for Implicit Image Segmentation Ensembles}

\midlauthor{\Name{Julia Wolleb\midljointauthortext{Contributed equally}} \Email{julia.wolleb@unibas.ch}\\
\Name{Robin Sandk\"uhler\midlotherjointauthor}\Email{robin.sandkuehler@unibas.ch}\\
\Name{Florentin Bieder}\Email{florentin.bieder@unibas.ch}   \\
\Name{Philippe Valmaggia}\Email{philippe.valmaggia@unibas.ch}\\
\Name{Philippe C. Cattin}\Email{philippe.cattin@unibas.ch}\\
\addr Department of Biomedical Engineering, University of Basel, Allschwil, Switzerland }

\begin{document}

\maketitle

\begin{abstract}
Diffusion models have shown impressive performance for generative modelling of images. In this paper, we present a novel semantic segmentation method based on diffusion models. By modifying the training and sampling scheme, we show that diffusion models can perform lesion segmentation of medical images. To generate an image specific segmentation, we train the model on the ground truth segmentation, and use the image as a prior during training and in every step during the sampling process. With the given stochastic sampling process, we can generate a distribution of segmentation masks. This property allows us to compute pixel-wise uncertainty maps of the segmentation, and allows an implicit ensemble of segmentations that increases the segmentation performance. We evaluate our method on the BRATS2020 dataset for brain tumor segmentation. Compared to state-of-the-art segmentation models, our approach yields good segmentation results and, additionally, detailed uncertainty maps.
\end{abstract}

\begin{keywords}
Diffusion models, segmentation, uncertainty estimation
\end{keywords}

\section{Introduction}\label{Intro}
Semantic segmentation is an important and well-explored area in medical image analysis \cite{biomedseg}. The automated segmentation of lesions in medical images with machine learning has shown good performances \cite{nnunet} and is ready for clinical application to support diagnosis \cite{trial}. In medical applications, it is of high interest to measure the uncertainty of a given prediction, especially when used for further treatments like radiation therapy. \\
In this work, we focus on the BRATS2020 brain tumor segmentation challenge \cite{brats1, brats2, brats3}. This dataset provides four different MR sequences for each patient (namely T1-weighted, T2-weighted, FLAIR and T1-weighted with contrast enhancement), as well as the pixel-wise ground truth segmentation. An exemplary image can be found in Appendix \ref{app:brats}.\\
We propose a novel segmentation method based on a Denoising Diffusion Probabilistic Model (DDPM) \cite{DDPM}, which can provide uncertainty maps of the produced segmentation mask. An overview of the workflow for an image of the BRATS2020 dataset is shown in Figure \ref{fig:overview}. 
We train a DDPM on the segmentation masks and add the original brain MR image as an image prior to induce the anatomical information. As sampling with DDPMs has a stochastic element in each sampling step, we can generate many different segmentation masks for the same input image and the same pretrained model. This ensemble of segmentations allows us to compute the pixel-wise variance maps, which visualizes the uncertainty of the generated segmentation. Moreover, the ensembling of the segmentations in a mean map boosts the segmentation performance.\\
We compare ourselves against state-of-the-art segmentation algorithms, and visually compare our variance map against common uncertainty maps.
The code will be publicly available at \url{https://github.com/JuliaWolleb/Diffusion-based-Segmentation}.
\begin{figure}[t]
	\floatconts
	{fig:overview}
	{\caption{Workflow for the implicit generation of segmentation ensembles and uncertainty maps with diffusion models. The input image consists of four different MR sequences. Going $n$ times through the sampling process of the diffusion model with different Gaussian noise, $n$ different segmentation masks are generated.}}
	{\includegraphics[width=0.9\linewidth]{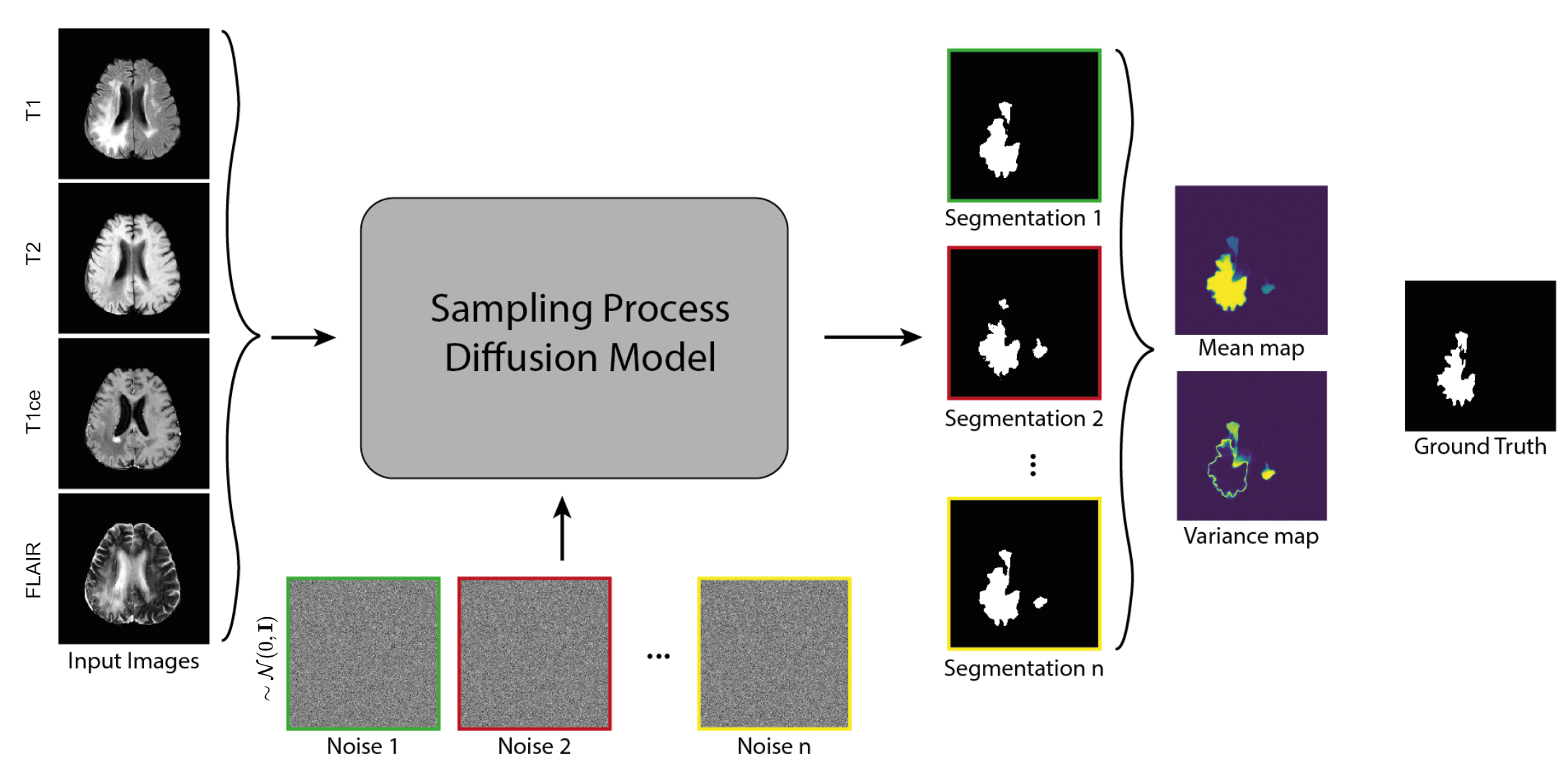}}
\end{figure}

\paragraph{Related Work}
In medical image segmentation, a common method is the application of a U-Net \cite{unet} or SegNet \cite{segnet} to predict the segmentation mask for every input image. 
This approach was successfully applied for many different tasks \cite{heart,MS,lung}. The state of the art is given by nnU-Nets \cite{nnunet}, where the best architecture and hyperparameters are automatically chosen for every specific dataset. \\
Uncertainty quantification is of high interest in deep learning research \cite{uncertainty}, which is often done using Bayesian neural networks \cite{bayesiansegnet, validity, galgar}. We can differentiate between epistemic uncertainty, which refers to uncertainty in the model parameters, and aleatoric uncertainty, which refers to uncertainty in the data. As stated in \cite{epivsaleo}, the epistemic uncertainty of a segmentation model can be approximated with Monte Carlo Dropout, whereas the aleatoric uncertainty can be modeled with MAP inference. Those methods were also applied on various medical tasks \cite{medbayes1,medbayes2,medbayes3}, including brain tumor segmentation\cite{bratsbayes1,jungo2,bratsbayes3}.\\
During the last year, DDPMs have gained a lot of attention due to their astonishing performance in image generation \cite{beatgans}. Images are generated by sampling from Gaussian noise. This sampling scheme follows a stochastic process, and therefore sampling from the same noisy image does not result in the same output image.
A different sampling scheme was introduced by Denoising Diffusion Implicit Models (DDIM)\cite{ddim}, where sampling is deterministic and can be done by skipping multiple steps. Moreover, meaningful interpolation between images can be achieved.
DDPM was further improved by \cite{improving} and \cite{beatgans}, where changes in the loss objective, architecture improvements, and classifier guidance during sampling improved the output image quality.\\
While some new work applies diffusion models on tasks such as image-to-image translation \cite{unitddpm}, style transfer \cite{ilvr}, or inpainting tasks \cite{palette}, so far there is only very little work about semantic segmentation.
Recently, one approach to perform semantic segmentation with a diffusion model was proposed by \cite{diffseg}. A DDPM is trained to reconstruct the image that should be segmented. Then, an MLP for classification is applied on the features of the model, which results in a segmentation mask for the original image. In contrast to this method, we train a DDPM directly to generate the segmentation mask.\\
Simultaneously and independent from us, \cite{segdiff} developed an image segmentation method similar to ours. However, they use a separate encoder for the image and the segmentation. Training a larger model may be difficult for medical image analysis due to possible large input images such as 3D data. Our method uses only one encoder to encode the image information and the segmentation mask.

\section{Method} \label{method}
The goal is to train a DDPM to generate segmentation masks. We follow the idea and implementation proposed in \cite{improving}.
The core idea of diffusion models is that for many timesteps $T$, noise is added to an image $x$. This results in a series of noisy images ${x_0, x_1, ..., x_T}$, where the noise level is steadily increased from $0$ (no noise) to $T$ (maximum noise). The model follows the architecture of a U-Net and predicts $x_{t-1}$ from $x_t$ for any step $t \in \{1,...,T\}$. During training, we know the ground truth for $x_{t-1}$, and the model is trained with an MSE loss. During sampling, we start from noise $x_T \sim \mathcal{N}(0,\mathbf{I})$, sample for $T$ steps, until we get a fake image  $x_{0}$.\\ 
The complete derivations of the formulas below can be found in \cite{DDPM, improving}.  
The main components of diffusion models are the forward noising process $q$ and the reverse denoising process $p$. Following \cite{DDPM}, the forward noising process $q$ for a given image $x$ at step $t$ is given by 
\begin{equation}\label{eq:forward}
q(x_{t}|x_{t-1}):=\mathcal{N}(x_{t};\sqrt{1-\beta _{t}}x_{t-1},\beta _{t}\mathbf{I}),
\end{equation}
where $\mathbf{I}$ denotes the identity matrix and $\beta_{1},...,\beta_{T}$ are the forward process variances. The idea is that in every step, a small amount of Gaussian noise is added to the image. Doing this for $t$ steps, we can write
\begin{equation}\label{eq:property_derivation}
q(x_{t}|x_{0}):=\mathcal{N}(x_{t};\sqrt{\overline{\alpha}_{t}}x_{0},(1-\overline{\alpha}_{t})\mathbf{I}),
\end{equation}
with $\alpha _{t}:=1-\beta _{t}$ and $\overline{\alpha}_{t}:=\prod_{s=1}^t \alpha _{s}$. 
With the reparametrization trick, we can directly write $x_t$ as a function of $x_0$:
\begin{equation}\label{eq:property}
x_{t}=\sqrt[]{\overline{\alpha} _{t}}x_{0}+\sqrt[]{1-\overline{\alpha} _{t}}\epsilon, \quad \mbox{with } \epsilon \sim \mathcal{N}(0,\mathbf{I}).
\end{equation}
The reverse process $p_{\theta}$ is learned by the model parameters $\theta$ and is given by 
\begin{equation}\label{eq:reverse}
p_{\theta}(x_{t-1}\vert x_t)\sim \mathcal{N}\bigl(x_{t-1};\mu_{\theta}(x_t, t), \Sigma_\theta(x_t,t)\bigr).
\end{equation}
As shown in \cite{DDPM}, we can then predict $x_{t-1}$ from $x_t$ with
\begin{equation}\label{eq:sampling}
x_{t-1}=\frac{1}{\sqrt[]{\alpha _{t}}}\left(x_{t}-\frac{1-\alpha _{t}}{\sqrt[]{1-\overline{\alpha }_{t}}}\epsilon _{\theta }(x_{t},t)\right)+\sigma_{t}\mathbf{z}, \quad \mbox{with } \mathbf{z} \sim \mathcal{N}(0,\mathbf{I}),
\end{equation}
where $\sigma_t$ denotes the variance scheme that can be learned by the model, as proposed in \cite{improving}.
We can see in Equation \ref{eq:sampling} that sampling has a random component $\mathbf{z}$, which leads to a stochastic sampling process.
Note that $\epsilon_{\theta}$ is the U-Net we train, with input ${x_{t}=\sqrt[]{\overline{\alpha} _{t}}x_{0}+\sqrt[]{1-\overline{\alpha} _{t}}\epsilon}$. The noise scheme $\epsilon _{\theta }(x_{t},t)$  that will be subtracted from $x_t$ during sampling according to Equation \ref{eq:sampling} has to be learned by the model. This U-Net is trained with the loss objectives given in \cite{improving}.\\
We now modify this idea to use diffusion models for semantic segmentation. A visualization of the workflow is given in Figure \ref{fig:example} for the task of brain tumor segmentation.\\
\begin{figure}[h]
	\floatconts
	{fig:example}
	{\caption{The training and sampling procedure of our method. In every step $t$, the anatomical information is induced by concatenating the brain MR images $b$ to the noisy segmentation mask $x_{b,t}$.}}
	{\includegraphics[width=0.9\linewidth]{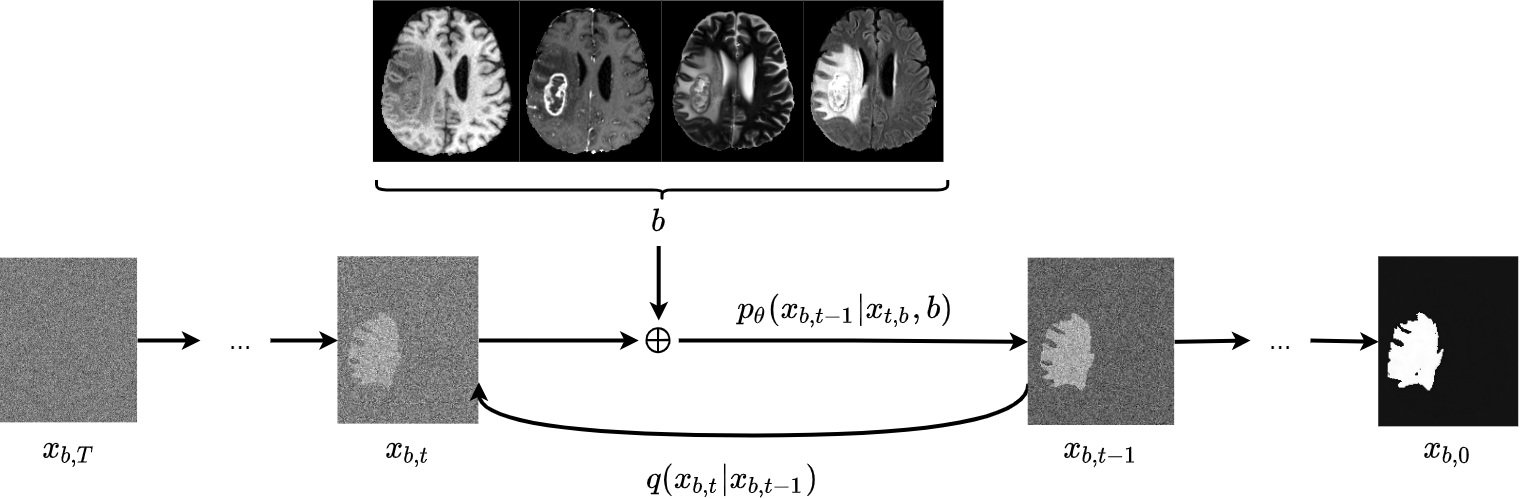}}
\end{figure}\\
 Let $b$ be the given brain MR image of dimension $(c,h,w)$, where $c$ denotes the number of channels, and $(h,w)$ denote the image height and image width. The ground truth segmentation of the tumor for the input image $b$ is denoted as $x_b$, and is of dimension $(1,h,w)$.
We train a DDPM for the generation of segmentation masks. In the classical DDPM approach, $x_b$ would be the only input we need for training, which would result in an arbitrary segmentation mask $x_0$ when we sample from noise during inference. In contrast to that, the goal in our proposed method is not to generate \textit{any} segmentation mask, but we want a meaningful segmentation mask $x_{b,0}$ for a given image $b$. To achieve this, we add additional channels to the input:
 We induce the anatomical information present in $b$ by adding it as an image prior to $x_b$.
We do this by concatenating $b$ and $x_b$, and define $X:=b \oplus x_b$.
Consequently, $X$ has dimension $(c+1,h,w)$.\\
During the noising process $q$, we only add noise to the ground truth segmentation $x_b$:

\begin{equation}
x_{b,t}=\sqrt[]{\overline{\alpha} _{t}}x_{b}+\sqrt[]{1-\overline{\alpha} _{t}}\epsilon, \quad \mbox{with } \epsilon \sim \mathcal{N}(0,\mathbf{I}), 
\end{equation}
and we define $X_t:=b \oplus x_{b,t}$.
Equation \ref{eq:sampling} is then altered to
\begin{equation}\label{eq:sampling2}
x_{b,t-1}=\frac{1}{\sqrt[]{\alpha _{t}}}\left(x_{b,t}-\frac{1-\alpha _{t}}{\sqrt[]{1-\overline{\alpha }_{t}}}\epsilon _{\theta }(X_{t},t)\right)+\sigma_{t}\mathbf{z}, \quad \mbox{with } \mathbf{z} \sim \mathcal{N}(0,\mathbf{I})
\end{equation}
and results in a slightly denoised $x_{b,t-1}$ of dimension $(1,h,w)$.
During inference, we follow the procedure presented in Algorithm \ref{alg:net}, which is a stochastic process.  Therefore, sampling twice for the same brain MR image $b$ does not result in the same segmentation mask prediction $x_{b,0}$.
Exploiting this property, we can implicitly generate an ensemble of segmentation masks without having to train a new model. This ensemble can then be used to boost the segmentation performance.

\begin{algorithm2e}
	\caption{Sampling Procedure}
	\label{alg:net}
	\KwIn{$b$, the original brain MRI}
	\KwOut{$x_{b,0}$, the predicted segmentation mask}
	sample $x_{b,T}\sim N(0,\mathbf{I})$\;
	\For{$t\leftarrow T$ \KwTo $1$}{
		$X_t \leftarrow b\oplus x_{b,t}$\;
	$	x_{b,t-1} \leftarrow \frac{1}{\sqrt[]{\alpha _{t}}}\left(x_{b,t}-\frac{1-\alpha _{t}}{\sqrt[]{1-\overline{\alpha }_{t}}}\epsilon _{\theta }(X_{t},t)\right)+\sigma_{t}\mathbf{z}, \quad \mbox{with } \mathbf{z} \sim \mathcal{N}(0,\mathbf{I})$ \;
	}
\end{algorithm2e}

\section{Dataset and Training Details}
We evaluate our method on the BRATS2020 dataset. As described in Section \ref{Intro}, images of four different MR sequences are provided for each patient, which are stacked to 4 channels.  We slice the 3D MR scans in axial slices. Since tumors rarely occur on the upper or lower part of the brain, we exclude the lowest 80 slices and the uppermost 26 slices.  For intensity normalization, we cut the top and bottom one percentile of the pixel intensities. We crop the images to a size of (4, 224, 224). The provided ground truth labels contain four classes, which are background,  GD-enhancing tumor, the peritumoral edema, and the necrotic and non-enhancing tumor core. We merge the three different tumor classes into one class and therefore define the segmentation problem as a pixel-wise binary classification. Our training set includes 16,298 images originating from 332 patients, and the test set comprises 1,082 images with non-empty ground truth segmentations, originating from 37 patients. No data augmentation is applied.\\
The hyperparameters for our DDPM models are described in the appendix of \cite{improving}. We choose a linear noise schedule for $T=1000$ steps. The model is trained with the hybrid loss objective, with a learning rate of ${10^{-4}}$ for the Adam optimizer, and a batch size of 10. The number of channels in the first layer is chosen as 128, and we use one attention head at resolution 16. We train the model for 60,000 iterations on an NVIDIA Quadro RTX 6000 GPU, which takes around one day.
The training details for the comparing methods can be found in Appendix \ref{implementation}.

\section{Results and Discussion}
During evaluation, we take an image $b$ from the test set, follow Algorithm \ref{alg:net} and produce a segmentation mask. This mask is thresholded at $0.5$ to obtain a binary segmentation. In Table \ref{tab:example},  the Dice score, the Jaccard index, and the 95 percentile Hausdorff Distance (HD95) are presented. We achieve good results with respect to all those metrics. \\
 For every image of the test set, we sample 5 different segmentation masks. This implicitly defines an ensemble by averaging over the 5 masks and thresholding it at 0.5. We report the results for this ensemble in the second line of Table \ref{tab:example}. We see that already an ensemble of 5 increases the performance of our approach.\\
 In the last column of Table \ref{tab:example}, we count the cases where the model produces an empty segmentation mask. This results in a Dice of zero, and HD95 cannot be computed. If we disregard those cases, we report the HD95 score, and the average Dice score and Jaccard index are reported in square brackets in Table \ref{tab:example}. \\
 As baseline, we report the segmentation scores for the nnU-Net and SegNet. By default, nnU-Net is an ensemble of a 5-fold cross validation. We also implement Bayesian SegNet with Monte Carlo dropout as proposed in \cite{bayesiansegnet}. By sampling five times during inference, we can again make an ensemble of the generated segmentation masks. The scores for this ensemble are reported in the last line of Table \ref{tab:example}.\\ The generation of one sample with our method takes 48 seconds, while the computation of the segmentation mask with SegNet takes 13 ms.  To speed up the sampling process, we will consider sampling with the DDIM approach in future work.
 \begin{table}[b]
	\floatconts
	{tab:example}%
	{\caption{Segmentation scores of our method and nnU-Net on different metrics.}}%
	{\begin{tabular}{lllll}
			\bfseries Method & \bfseries Dice  & \bfseries HD95& \bfseries Jaccard& \bfseries  empty\\
			Ours (1 sampling run) & 0.866 [0.892]&6.052&0.795 [0.819]&31\\
		 Ours  (ensemble of 5 runs)  &0.881 [0.909]& 5.178 & 0.819 [0.845]&34\\
			nnU-Net (ensemble of 5-fold cross-val.)  & 0.891 [0.905]&5.004&0.831 [0.845]&17\\
			SegNet (1 run)  &0.839 [0.867]&7.190&0.761 [0.786]&34\\
			Bayesian SegNet (ensemble of 5 runs) &0.838 [0.841]&13.707&0.747 [0.749]&3
	\end{tabular}}
\end{table}\\
For visualization of the uncertainty maps, we select three exemplary images $b_1$, $b_2$, and $b_3$ from the test set. More examples are presented in Appendix \ref{examples}.
 To generate detailed uncertainty maps, we sample 100 segmentation masks for each of the images, and compute the pixel-wise variance.
In Figure \ref{varmap}, we present one channel of the original brain MR image $b$, the ground truth segmentation, two different sampled segmentation masks, as well as the mean and variance map. We can clearly identify the areas where the model was uncertain. Moreover, by thresholding the mean map at $0.5$, we can produce the ensembled segmentation mask.
In Table \ref{tab:ens}, we report the segmentation scores and  for this ensemble mask, as well as the average scores for the 100 samples. We see that the ensemble can boost the performance for the examples $b_1$, $b_2$ and $b_3$.\\
\begin{table}[htbp]
	\floatconts
	{tab:ens}%
	{\caption{Segmentation scores for the 100 samples of the examples presented in Figure \ref{varmap}. }}%
	{\begin{tabular}{c|cSc|cSc|}
&  \multicolumn{3}{c|}{\textbf{Average}} & \multicolumn{3}{c|}{\textbf{Ensemble}} \\
			\bfseries Example & \textbf{Dice}  & \textbf{HD95} & \textbf{Jaccard}&\textbf{Dice}& \textbf{ HD95}  & \textbf{Jaccard}\\
			$b_1$&0.969&2.360&0.939&0.981&1.000&0.962 \\
			$b_2$ &0.869&18.503&0.769&0.885&18.468&0.783\\
			$b_3$ & 0.932&5.227 &0.872&0.952&4.474&0.907\\
	\end{tabular}}
\end{table}
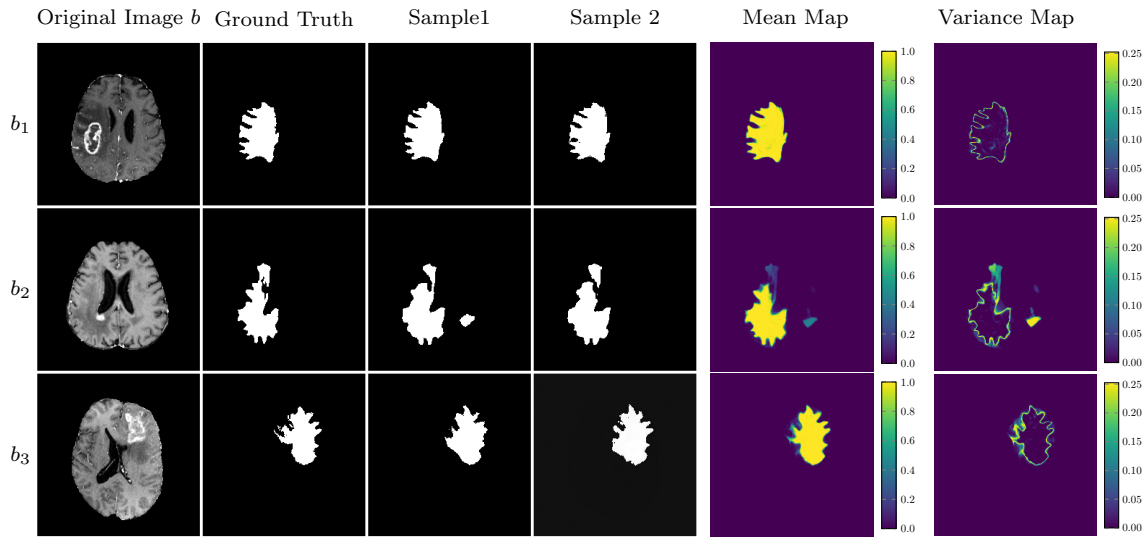
\begin{figure}[t]
	\label{varmap}
	\begin{tikzpicture}
	\node[draw=black, inner sep=0pt, thick] at (0, 0) {\includegraphics[scale=0.27]{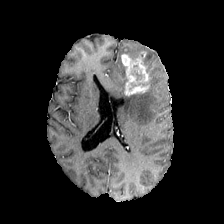}};
	\node[] at (0,5.8)  {\scriptsize Original Image $b$};
	\node[draw=black, inner sep=0pt, thick] at (2.2, 0) {\includegraphics[scale=0.27]{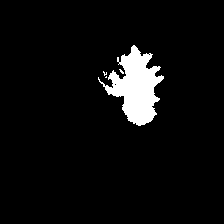}};
	\node[] at (2.2, 5.8)  {\scriptsize Ground Truth};
	\node[draw=black, inner sep=0pt, thick] at (4.4, 0) {\includegraphics[scale=0.27]{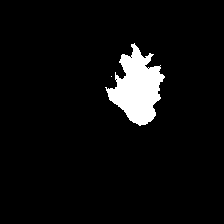}};
	\node[] at (4.4, 5.8) {\scriptsize Sample1};
	\node[draw=black, inner sep=0pt, thick] at (6.6, 0) {\includegraphics[scale=0.27]{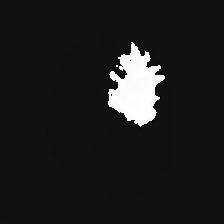}};
\node[] at (6.6, 5.8) {\scriptsize Sample 2};
		\node[draw=white, inner sep=0pt, thick] at (9.25, 0) {\resizebox{0.182\textwidth}{!}{\input{mean000246.tex}}};
	\node[] at (9, 5.8) {\scriptsize Mean Map};
	\node[draw=white, inner sep=0pt, thick] at (12.25, 0) {\resizebox{0.184\textwidth}{!}{\input{var000246.tex}}};
	\node[] at (11.8, 5.8) {\scriptsize Variance Map};
	
	\node[draw=black, inner sep=0pt, thick] at (0,  2.2) {\includegraphics[scale=0.27]{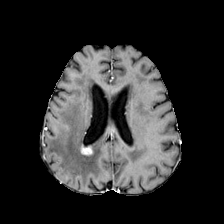}};
	\node[draw=black, inner sep=0pt, thick] at (2.2, 2.2) {\includegraphics[scale=0.27]{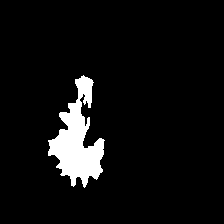}};
	\node[draw=black, inner sep=0pt, thick] at (4.4,  2.2) {\includegraphics[scale=0.27]{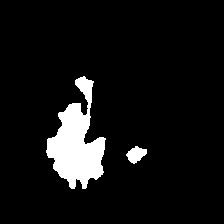}};
	\node[draw=black, inner sep=0pt, thick] at (6.6,  2.2) {\includegraphics[scale=0.27]{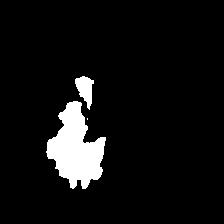}};
	\node[draw=white, inner sep=0pt, thick] at (9.25,  2.2) {\resizebox{0.182\textwidth}{!}{\input{mean016867}}};
	\node[draw=white, inner sep=0pt, thick] at (12.25,  2.2){\resizebox{0.184\textwidth}{!}{\input{var016867}}};


	\node[draw=black, inner sep=0pt, thick] at (0, 4.4) {\includegraphics[scale=0.27]{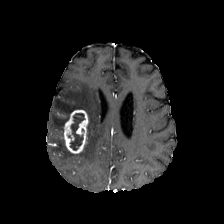}};
	\node[draw=black, inner sep=0pt, thick] at (2.2, 4.4) {\includegraphics[scale=0.27]{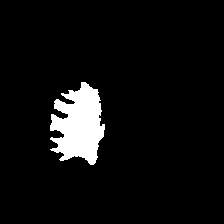}};
	\node[draw=black, inner sep=0pt, thick] at (4.4,4.4) {\includegraphics[scale=0.27]{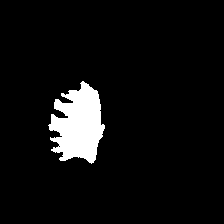}};
		\node[draw=black, inner sep=0pt, thick] at (6.6, 4.4) {\includegraphics[scale=0.27]{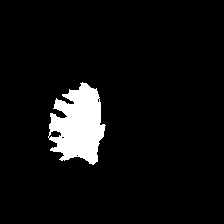}};
		\node[draw=white, inner sep=0pt, thick] at (9.25, 4.4) {\resizebox{0.182\textwidth}{!}{\input{mean002468}}};
	\node[draw=white, inner sep=0pt, thick] at (12.25, 4.4) {\resizebox{0.184\textwidth}{!}{\input{var002468.tex}}};

	\node at (-1.3,0) {\scriptsize $b_3$};
	\node at (-1.3,2.2) {\scriptsize $b_2$};
	\node at (-1.3,4.4) {\scriptsize $b_1$};

	\end{tikzpicture}
	\vspace{2pt}
	\caption{ Examples of the produced mean and variance maps for 100 sampling runs. } \label{varmap}
\end{figure}\\
In Figure \ref{fig:plot}, we plot the number of samples in the ensemble against the Dice score for the three examples $b_1$, $b_2$, and $b_3$. We can see that already an ensemble of five samples improves the performance, and then the curve flattens. In \cite{segdiff}, a similar experiment was performed on a different data set. Independently from each other, we got the same findings. 
In Figure \ref{varmapsvergleich}, we compare our variance maps against the ones of the Bayesian SegNet with Monte Carlo (MC) dropout for 100 samples, as well as the aleatoric uncertainty maps for SegNet, computed as proposed in \cite{epivsaleo}. 

\begin{figure}[h]
	\floatconts
	{fig:plot}
	{\caption{Performance of the ensemble with respect to the number of samples for the examples $b_1$, $b_2$, and $b_3$,  presented in Figure \ref{varmap}.}}
{\resizebox{0.38\textwidth}{!}{\input{plot.tex}}}
\end{figure}
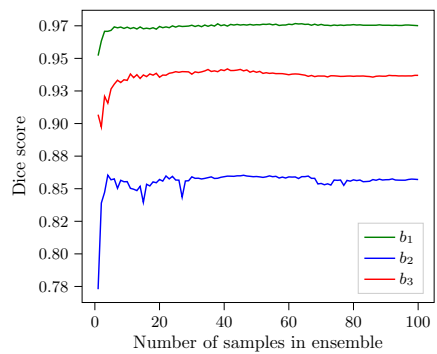

\begin{figure}[t]
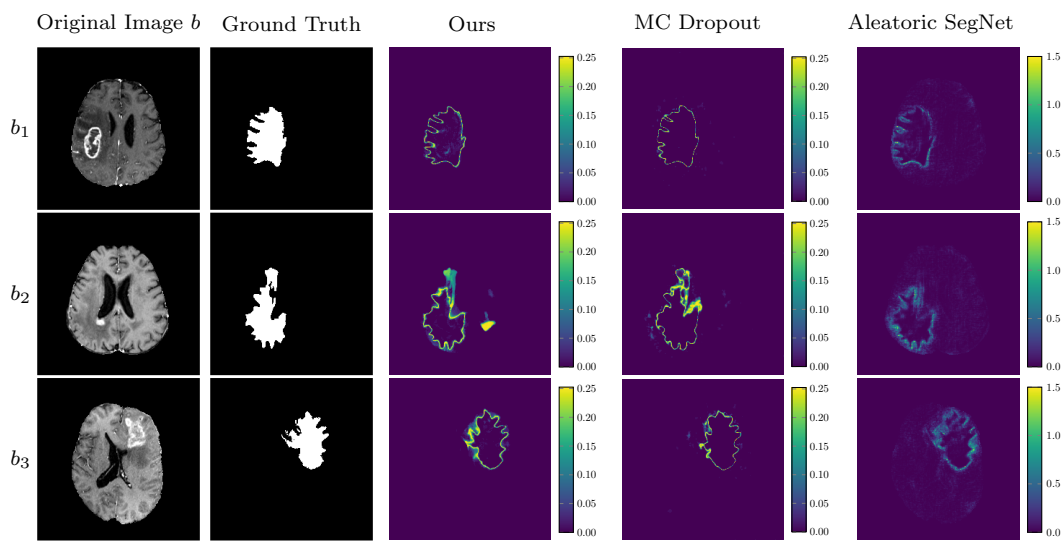

	\floatconts
	{varmapsvergleich}
	{\caption{Comparison of the different uncertainty maps for the three examples.}}
	{\begin{tikzpicture}
	\node[draw=black, inner sep=0pt, thick] at (0, 0) {\includegraphics[scale=0.27]{000246/brain}};
	\node[] at (0,5.8)  {\scriptsize Original Image $b$};
	\node[draw=black, inner sep=0pt, thick] at (2.3, 0) {\includegraphics[scale=0.27]{000246/gt}};
	\node[] at (2.3, 5.8)  {\scriptsize Ground Truth};
	\node[draw=white, inner sep=0pt, thick] at (5, 0) 
	{\resizebox{0.184\textwidth}{!}{\input{var000246.tex}}};
	\node[] at (4.7, 5.8) {\scriptsize Ours};
	\node[draw=white, inner sep=0pt, thick] at (8.1, 0) 
	{\resizebox{0.183\textwidth}{!}{\input{bayes000246}}};
\node[] at (7.7, 5.8) {\scriptsize MC Dropout};
		\node[draw=white, inner sep=0pt, thick] at (11.2, 0)
		{\resizebox{0.18\textwidth}{!}{\input{alea_000246}}};
	\node[] at (10.8, 5.8) {\scriptsize Aleatoric SegNet};

	\node[draw=black, inner sep=0pt, thick] at (0,  2.2) {\includegraphics[scale=0.27]{016867/brain}};
	\node[draw=black, inner sep=0pt, thick] at (2.3, 2.2) {\includegraphics[scale=0.27]{016867/gt}};
	\node[draw=white, inner sep=0pt, thick] at (5,  2.2) {\resizebox{0.184\textwidth}{!}{\input{var016867.tex}}};
	\node[draw=white, inner sep=0pt, thick] at (8.1,  2.2)
	{\resizebox{0.183\textwidth}{!}{\input{bayes016867}}};
	\node[draw=white, inner sep=0pt, thick] at (11.2,  2.2)
	{\resizebox{0.18\textwidth}{!}{\input{alea_016867}}};
	\node[draw=black, inner sep=0pt, thick] at (0, 4.4) {\includegraphics[scale=0.27]{002468/brain}};
	\node[draw=black, inner sep=0pt, thick] at (2.3, 4.4) {\includegraphics[scale=0.27]{002468/gt}};
	\node[draw=white, inner sep=0pt, thick] at (5, 4.4) 
	 {\resizebox{0.184\textwidth}{!}{\input{var002468.tex}}};
		\node[draw=white, inner sep=0pt, thick] at (8.1, 4.4)
		{\resizebox{0.183\textwidth}{!}{\input{bayes002468}}};
		\node[draw=white, inner sep=0pt, thick] at (11.2,4.4) 
			 {\resizebox{0.18\textwidth}{!}{\input{alea_002468}}};

	\node at (-1.3,0) {\scriptsize $b_3$};
	\node at (-1.3,2.2) {\scriptsize $b_2$};
	\node at (-1.3,4.4) {\scriptsize $b_1$};

	\end{tikzpicture}} 	
\end{figure}

\section{Conclusion }
We presented a novel approach for biomedical image segmentation based on DDPMs.
Using the stochastic sampling process, our method allows implicit ensembling of different segmentation masks for the same input brain MR image, without having to train a new model. We could show that ensembling those segmentation masks increases the performance of the model with respect to different segmentation scores. Moreover,  we can generate uncertainty maps by computing the variance of the different segmentation masks. This is of great interest in clinical applications, when we want to measure the uncertainty of the decision of the model.
For future work, we plan to investigate the segmentation of the different tumor classes provided by the BRATS2020 challenge.

\midlacknowledgments{This research was supported by the Novartis FreeNovation initiative and the Uniscientia
Foundation (project \# 147-2018).}

\bibliography{midl-samplebibliography}

\appendix

\section{Exemplary Image of BRATS2020}\label{app:brats}

\begin{figure}[h]
	\floatconts
	{fig:bratsexample}
	{\caption{Exemplary image of the BRATS2020 dataset, with four different MR sequences and the ground truth segmentation.}}
	{\includegraphics[width=0.62\linewidth]{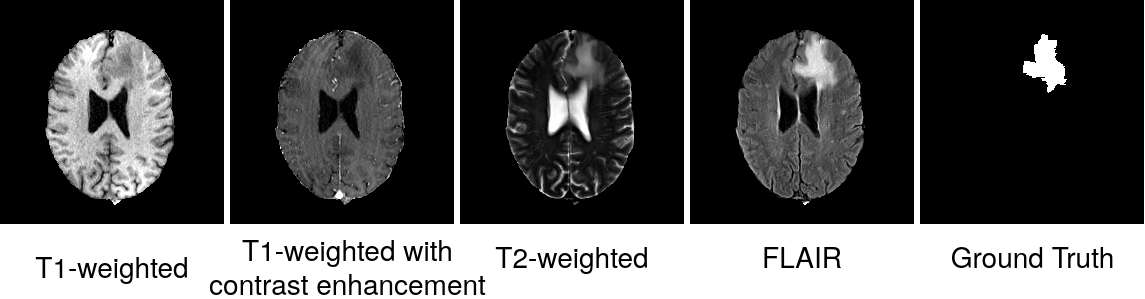}}
\end{figure}

\section{Implementation Details}\label{implementation}

We provide implementation details of the comparing methods.
\begin{itemize}
\item SegNet: We train the SegNet as proposed in \cite{segnet},  with a learning rate of ${10^{-4}}$ for the Adam optimizer and a batch size of 20. Training is performed with the binary cross-entropy loss and is stopped after 100 epochs.
\item Bayesian SegNet: We adapt the SegNet architecture, and place the dropout layers with a dropout probability of $p=0.5$ as proposed in \cite{bayesianseg}. The training schedule is kept the same as for SegNet.
\item nnU-Net: We take over all hyperparameter settings as proposed in their official implementation, which can be found at \url{https://github.com/MIC-DKFZ/nnUNet}.
\item Aleatoric Uncertainty Estimation: We keep the training settings for SegNet. The only change we need to make to the SegNet architecture is to double the number of output channels, such that we get both a prediction and a variance map. We follow the aleatoric loss implementation as proposed in \cite{jungo2}, which can be found at \url{https://github.com/alainjungo/reliability-challenges-uncertainty}.
\end{itemize}

\section{Further Examples}\label{examples}

We provide the mean and variance maps of three more exemplary images $b_4$, $b_5$, and $b_6$ of the test set.

\begin{figure}[h]
	\begin{tikzpicture}
	\node[draw=black, inner sep=0pt, thick] at (0, 0) {\includegraphics[scale=0.27]{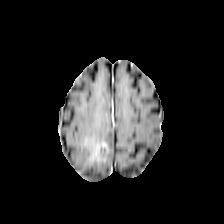}};
	\node[] at (0,5.8)  {\scriptsize Original Image $b$};
	\node[draw=black, inner sep=0pt, thick] at (2.2, 0) {\includegraphics[scale=0.27]{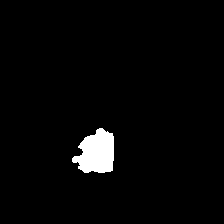}};
	\node[] at (2.2, 5.8)  {\scriptsize Ground Truth};
	\node[draw=black, inner sep=0pt, thick] at (4.4, 0) {\includegraphics[scale=0.27]{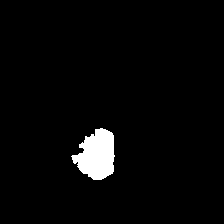}};
	\node[] at (4.4, 5.8) {\scriptsize Sample1};
	\node[draw=black, inner sep=0pt, thick] at (6.6, 0) {\includegraphics[scale=0.27]{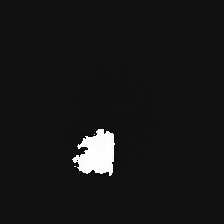}};
\node[] at (6.6, 5.8) {\scriptsize Sample 2};
		\node[draw=white, inner sep=0pt, thick] at (9.2, 0) 
		{\resizebox{0.182\textwidth}{!}{\input{mean013856}}};
	\node[] at (9, 5.8) {\scriptsize Mean Map};
	\node[draw=white, inner sep=0pt, thick] at (12.2, 0) 
	{\resizebox{0.183\textwidth}{!}{\input{var013856}}};
	\node[] at (11.8, 5.8) {\scriptsize Variance Map};
	
	\node[draw=black, inner sep=0pt, thick] at (0,  2.2) {\includegraphics[scale=0.27]{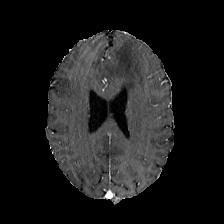}};
	\node[draw=black, inner sep=0pt, thick] at (2.2, 2.2) {\includegraphics[scale=0.27]{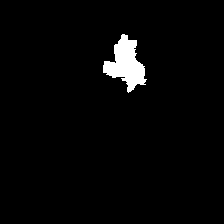}};
	\node[draw=black, inner sep=0pt, thick] at (4.4,  2.2) {\includegraphics[scale=0.27]{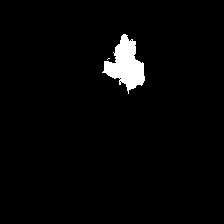}};
	\node[draw=black, inner sep=0pt, thick] at (6.6,  2.2) {\includegraphics[scale=0.27]{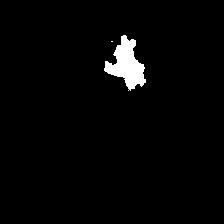}};
	\node[draw=white, inner sep=0pt, thick] at (9.2,  2.2)
	{\resizebox{0.182\textwidth}{!}{\input{mean001919}}};
	\node[draw=white, inner sep=0pt, thick] at (12.2,  2.2) 
	{\resizebox{0.183\textwidth}{!}{\input{var001919}}};

	\node[draw=black, inner sep=0pt, thick] at (0, 4.4) {\includegraphics[scale=0.27]{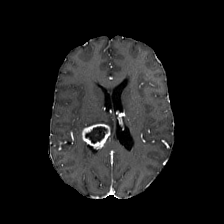}};
	\node[draw=black, inner sep=0pt, thick] at (2.2, 4.4) {\includegraphics[scale=0.27]{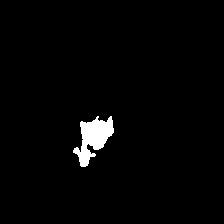}};
	\node[draw=black, inner sep=0pt, thick] at (4.4,4.4) {\includegraphics[scale=0.27]{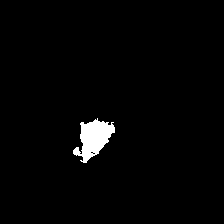}};
		\node[draw=black, inner sep=0pt, thick] at (6.6, 4.4) {\includegraphics[scale=0.27]{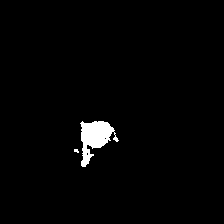}};
		\node[draw=white, inner sep=0pt, thick] at (9.2, 4.4)
		{\resizebox{0.182\textwidth}{!}{\input{mean004368}}};
	\node[draw=white, inner sep=0pt, thick] at (12.2, 4.4) 
	{\resizebox{0.183\textwidth}{!}{\input{var004368}}};

	\node at (-1.3,0) {\scriptsize $b_6$};
	\node at (-1.3,2.2) {\scriptsize $b_5$};
	\node at (-1.3,4.4) {\scriptsize $b_4$};

	\end{tikzpicture}
	\vspace{2pt}
	\caption{ Additional examples of the produced mean and variance maps for 100 sampling runs. } \label{morexamples}
\end{figure}
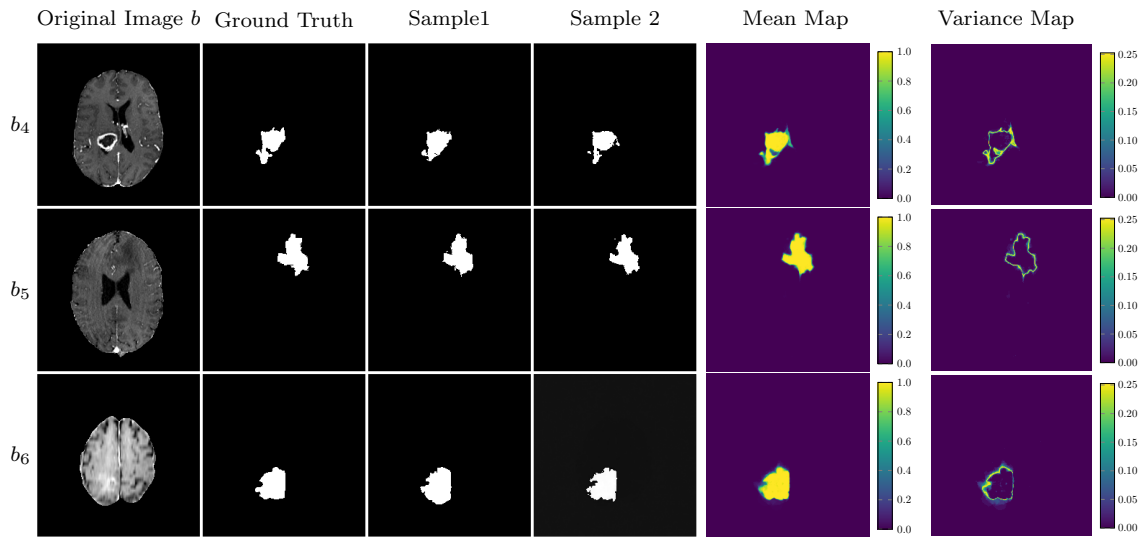

\end{document}

%% file: mean000246.tex
\begin{tikzpicture}

\begin{axis}[
hide axis,
axis equal image,
colorbar,
colorbar style={height=0.9*\pgfkeysvalueof{/pgfplots/parent axis width},ytick={0,0.2,0.4,0.6,0.8,1},yticklabels={0.0,0.2,0.4,0.6,0.8,1.0},ylabel={}, yshift=-3mm,yticklabel style={text width=1.7em,align=right}},
colormap/viridis,
point meta max=1,
point meta min=0,
tick align=outside,
tick pos=left,
x grid style={white!69.0196078431373!black},
xmin=-0.5, xmax=223.5,
xtick style={color=black},
y dir=reverse,
y grid style={white!69.0196078431373!black},
ymin=-0.5, ymax=223.5,
ytick style={color=black}
]
\addplot graphics [includegraphics cmd=\pgfimage,xmin=-0.5, xmax=223.5, ymin=223.5, ymax=-0.5] {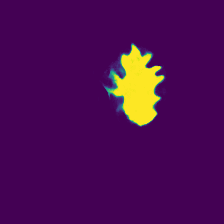};
\end{axis}

\end{tikzpicture}

%% file: var000246.tex
\begin{tikzpicture}

\begin{axis}[
hide axis,
axis equal image,
colorbar,
colorbar style={ytick={0,0.05,0.1,0.15,0.2,0.25},yticklabels={0.00,0.05,0.10,0.15,0.20,0.25},ylabel={},height=0.9*\pgfkeysvalueof{/pgfplots/parent axis width},yshift=-3mm,yticklabel style={text width=2.1em,align=right}},
colormap/viridis,
point meta max=0.252525240182877,
point meta min=0,
tick align=outside,
tick pos=left,
x grid style={white!69.0196078431373!black},
xmin=-0.5, xmax=223.5,
xtick style={color=black},
y dir=reverse,
y grid style={white!69.0196078431373!black},
ymin=-0.5, ymax=223.5,
ytick style={color=black}
]
\addplot graphics [includegraphics cmd=\pgfimage,xmin=-0.5, xmax=223.5, ymin=223.5, ymax=-0.5] {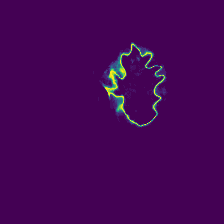};
\end{axis}

\end{tikzpicture}

%% file: mean016867.tex
\begin{tikzpicture}

\begin{axis}[
hide axis,
axis equal image,
colorbar,
colorbar style={height=0.9*\pgfkeysvalueof{/pgfplots/parent axis width},ytick={0,0.2,0.4,0.6,0.8,1},yticklabels={0.0,0.2,0.4,0.6,0.8,1.0},ylabel={}, yshift=-3mm,yticklabel style={text width=1.7em,align=right}},
colormap/viridis,
point meta max=1,
point meta min=0,
tick align=outside,
tick pos=left,
x grid style={white!69.0196078431373!black},
xmin=-0.5, xmax=223.5,
xtick style={color=black},
y dir=reverse,
y grid style={white!69.0196078431373!black},
ymin=-0.5, ymax=223.5,
ytick style={color=black}
]
\addplot graphics [includegraphics cmd=\pgfimage,xmin=-0.5, xmax=223.5, ymin=223.5, ymax=-0.5] {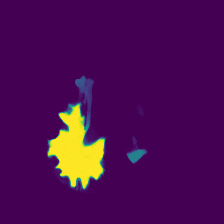};
\end{axis}

\end{tikzpicture}

%% file: var016867.tex
\begin{tikzpicture}

\begin{axis}[
hide axis,
axis equal image,
colorbar,
colorbar style={ytick={0,0.05,0.1,0.15,0.2,0.25},yticklabels={0.00,0.05,0.10,0.15,0.20,0.25},ylabel={},height=0.9*\pgfkeysvalueof{/pgfplots/parent axis width},yshift=-3mm,yticklabel style={text width=2.1em,align=right}},
colormap/viridis,
point meta max=0.252525240182877,
point meta min=0,
tick align=outside,
tick pos=left,
x grid style={white!69.0196078431373!black},
xmin=-0.5, xmax=223.5,
xtick style={color=black},
y dir=reverse,
y grid style={white!69.0196078431373!black},
ymin=-0.5, ymax=223.5,
ytick style={color=black}
]
\addplot graphics [includegraphics cmd=\pgfimage,xmin=-0.5, xmax=223.5, ymin=223.5, ymax=-0.5] {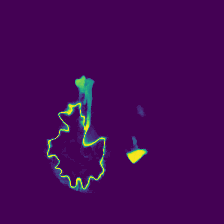};
\end{axis}

\end{tikzpicture}

%% file: mean002468.tex
\begin{tikzpicture}

\begin{axis}[
hide axis,
axis equal image,
colorbar,
colorbar style={height=0.9*\pgfkeysvalueof{/pgfplots/parent axis width},ytick={0,0.2,0.4,0.6,0.8,1},yticklabels={0.0,0.2,0.4,0.6,0.8,1.0},ylabel={}, yshift=-3mm,yticklabel style={text width=1.7em,align=right}},
colormap/viridis,
point meta max=1,
point meta min=0,
tick align=outside,
tick pos=left,
x grid style={white!69.0196078431373!black},
xmin=-0.5, xmax=223.5,
xtick style={color=black},
y dir=reverse,
y grid style={white!69.0196078431373!black},
ymin=-0.5, ymax=223.5,
ytick style={color=black}
]
\addplot graphics [includegraphics cmd=\pgfimage,xmin=-0.5, xmax=223.5, ymin=223.5, ymax=-0.5] {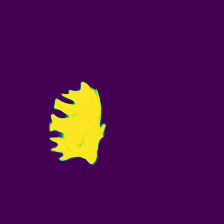};
\end{axis}

\end{tikzpicture}

%% file: var002468.tex
\begin{tikzpicture}

\begin{axis}[
hide axis,
axis equal image,
colorbar,
colorbar style={ytick={0,0.05,0.1,0.15,0.2,0.25},yticklabels={0.00,0.05,0.10,0.15,0.20,0.25},ylabel={},height=0.9*\pgfkeysvalueof{/pgfplots/parent axis width},yshift=-3mm,yticklabel style={text width=2.1em,align=right}},
colormap/viridis,
point meta max=0.252525240182877,
point meta min=0,
tick align=outside,
tick pos=left,
x grid style={white!69.0196078431373!black},
xmin=-0.5, xmax=223.5,
xtick style={color=black},
y dir=reverse,
y grid style={white!69.0196078431373!black},
ymin=-0.5, ymax=223.5,
ytick style={color=black}
]
\addplot graphics [includegraphics cmd=\pgfimage,xmin=-0.5, xmax=223.5, ymin=223.5, ymax=-0.5] {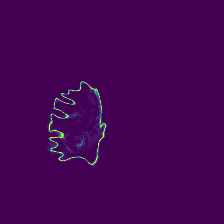};
\end{axis}

\end{tikzpicture}

%% file: plot.tex
\begin{tikzpicture}
\footnotesize
\begin{axis}[
legend cell align={left},
legend style={fill opacity=0.8, draw opacity=1, text opacity=1, at={(0.97,0.03)}, anchor=south east, draw=white!80!black},
tick align=outside,
tick pos=left,
x grid style={white!69.0196078431373!black},
xlabel={\small Number of samples in ensemble},
xmin=-3.95, xmax=104.95,
xtick style={color=black},
xtick={-20,0,20,40,60,80,100,120},
xticklabels={\(\displaystyle {−20}\),\(\displaystyle {0}\),\(\displaystyle {20}\),\(\displaystyle {40}\),\(\displaystyle {60}\),\(\displaystyle {80}\),\(\displaystyle {100}\),\(\displaystyle {120}\)},
y grid style={white!69.0196078431373!black},
ylabel={\small Dice score},
ymin=0.762682765722275, ymax=0.986913472414017,
ytick style={color=black},
ytick={0.75,0.775,0.8,0.825,0.85,0.875,0.9,0.925,0.95,0.975,1},
yticklabels={0.75,0.78,0.80,0.82,0.85,0.88,0.90,0.93,0.95,0.97,1.00}
]
\addplot [thick, green!50!black]
table {%
1 0.952100694179535
2 0.963326036930084
3 0.970758318901062
4 0.970754384994507
5 0.971494495868683
6 0.97416365146637
7 0.973454236984253
8 0.97402811050415
9 0.972990751266479
10 0.973936021327972
11 0.972857356071472
12 0.973632156848907
13 0.972546458244324
14 0.974113762378693
15 0.972706317901611
16 0.973168730735779
17 0.972715318202972
18 0.973506689071655
19 0.972413778305054
20 0.974477171897888
21 0.973541498184204
22 0.974156379699707
23 0.973541498184204
24 0.974934041500092
25 0.974004626274109
26 0.974925756454468
27 0.974468767642975
28 0.974620938301086
29 0.974798202514648
30 0.975424706935883
31 0.974468767642975
32 0.975263833999634
33 0.974316775798798
34 0.975577533245087
35 0.974620938301086
36 0.975569486618042
37 0.974781632423401
38 0.976544439792633
39 0.97510302066803
40 0.97573858499527
41 0.974773287773132
42 0.976052820682526
43 0.975577533245087
44 0.976060748100281
45 0.975577533245087
46 0.975883722305298
47 0.975561439990997
48 0.976044952869415
49 0.975561439990997
50 0.975883722305298
51 0.975730538368225
52 0.976206183433533
53 0.976052820682526
54 0.976375341415405
55 0.975730538368225
56 0.976214051246643
57 0.9752556681633
58 0.975569486618042
59 0.975593686103821
60 0.976229786872864
61 0.975915551185608
62 0.976721167564392
63 0.976398766040802
64 0.976552188396454
65 0.976068675518036
66 0.976229786872864
67 0.975432813167572
68 0.976229786872864
69 0.975424706935883
70 0.976068675518036
71 0.975424706935883
72 0.975416600704193
73 0.97510302066803
74 0.975585639476776
75 0.975263833999634
76 0.975585639476776
77 0.975280165672302
78 0.975593686103821
79 0.975593686103821
80 0.975907564163208
81 0.975585639476776
82 0.97574657201767
83 0.975432813167572
84 0.975424706935883
85 0.975424706935883
86 0.975424706935883
87 0.975432813167572
88 0.97575455904007
89 0.975432813167572
90 0.975424706935883
91 0.975432813167572
92 0.975263833999634
93 0.975432813167572
94 0.975263833999634
95 0.975271999835968
96 0.975593686103821
97 0.975601732730865
98 0.975424706935883
99 0.975271999835968
100 0.975094854831696
};
\addlegendentry{$b_1$}
\addplot [thick, blue]
table {%
1 0.772875070571899
2 0.839006125926971
3 0.847185015678406
4 0.860381245613098
5 0.856851041316986
6 0.85753470659256
7 0.850337862968445
8 0.856460988521576
9 0.855343759059906
10 0.855431973934174
11 0.850297391414642
12 0.849608719348907
13 0.848546326160431
14 0.851832926273346
15 0.839551627635956
16 0.853396892547607
17 0.852147340774536
18 0.855242967605591
19 0.854576289653778
20 0.85709422826767
21 0.855840623378754
22 0.859759211540222
23 0.857528269290924
24 0.859464943408966
25 0.85666161775589
26 0.856562912464142
27 0.843489825725555
28 0.855935096740723
29 0.856132090091705
30 0.859123945236206
31 0.85796320438385
32 0.858061254024506
33 0.857481002807617
34 0.857916116714478
35 0.858492136001587
36 0.859073996543884
37 0.857865452766418
38 0.85801112651825
39 0.858779907226562
40 0.859214305877686
41 0.859643220901489
42 0.859595954418182
43 0.859498560428619
44 0.859888732433319
45 0.859885454177856
46 0.860320150852203
47 0.859740674495697
48 0.859501361846924
49 0.85921186208725
50 0.859359204769135
51 0.859069466590881
52 0.858686447143555
53 0.859116971492767
54 0.859841465950012
55 0.85921186208725
56 0.859504103660583
57 0.858201444149017
58 0.858924686908722
59 0.858201444149017
60 0.859069466590881
61 0.85892254114151
62 0.859019696712494
63 0.857815146446228
64 0.859259247779846
65 0.859259247779846
66 0.859693467617035
67 0.85795933008194
68 0.85795933008194
69 0.853654444217682
70 0.854034125804901
71 0.853034615516663
72 0.853748321533203
73 0.852798640727997
74 0.856663286685944
75 0.856424033641815
76 0.856663286685944
77 0.8525630235672
78 0.85637491941452
79 0.855799913406372
80 0.856806695461273
81 0.856232166290283
82 0.856662750244141
83 0.85536915063858
84 0.855512678623199
85 0.855706393718719
86 0.856998980045319
87 0.856472194194794
88 0.857334673404694
89 0.856663882732391
90 0.856903195381165
91 0.856951177120209
92 0.857526421546936
93 0.856663882732391
94 0.856567680835724
95 0.856520295143127
96 0.85651957988739
97 0.857094943523407
98 0.857382535934448
99 0.857238709926605
100 0.857046961784363
};
\addlegendentry{$b_2$}
\addplot [thick, red]
table {%
1 0.906821966171265
2 0.897397935390472
3 0.9208163022995
4 0.915667653083801
5 0.926524043083191
6 0.930223107337952
7 0.933226048946381
8 0.931467950344086
9 0.93360161781311
10 0.933117806911469
11 0.938014030456543
12 0.935216963291168
13 0.937487423419952
14 0.934598326683044
15 0.937273800373077
16 0.936024188995361
17 0.938077986240387
18 0.937122106552124
19 0.938554227352142
20 0.935535848140717
21 0.937186419963837
22 0.937085449695587
23 0.938563108444214
24 0.939004838466644
25 0.939787983894348
26 0.939315021038055
27 0.939836084842682
28 0.939763844013214
29 0.939648270606995
30 0.937875747680664
31 0.939648270606995
32 0.939126968383789
33 0.939648270606995
34 0.940376162528992
35 0.941481947898865
36 0.939951956272125
37 0.940423846244812
38 0.939787983894348
39 0.941270470619202
40 0.940635621547699
41 0.94192773103714
42 0.94065934419632
43 0.940706729888916
44 0.941011786460876
45 0.941528618335724
46 0.940423846244812
47 0.941106021404266
48 0.939387857913971
49 0.940259754657745
50 0.939315021038055
51 0.940024018287659
52 0.938489258289337
53 0.938987791538239
54 0.938489258289337
55 0.938726484775543
56 0.938301265239716
57 0.938326001167297
58 0.938276529312134
59 0.93790066242218
60 0.937449872493744
61 0.937925517559052
62 0.938063740730286
63 0.937925517559052
64 0.9375
65 0.93712455034256
66 0.936221420764923
67 0.936886370182037
68 0.936008036136627
69 0.936485648155212
70 0.936434745788574
71 0.936434745788574
72 0.935768783092499
73 0.93667334318161
74 0.936622560024261
75 0.936434745788574
76 0.936170220375061
77 0.936221420764923
78 0.93640923500061
79 0.936460196971893
80 0.936221420764923
81 0.936434745788574
82 0.93640923500061
83 0.936485648155212
84 0.936221420764923
85 0.936059355735779
86 0.935632646083832
87 0.93627256155014
88 0.936460196971893
89 0.936323583126068
90 0.936323583126068
91 0.936962187290192
92 0.936749398708344
93 0.936749398708344
94 0.936511099338531
95 0.936561942100525
96 0.936511099338531
97 0.936561942100525
98 0.936536550521851
99 0.936987400054932
100 0.936987400054932
};
\addlegendentry{$b_3$}
\end{axis}

\end{tikzpicture}

%% file: bayes000246.tex
\begin{tikzpicture}

\begin{axis}[
hide axis,
axis equal image,
colorbar,
colorbar style={height=0.9*\pgfkeysvalueof{/pgfplots/parent axis width},yshift=-3mm, ytick={0,0.05,0.1,0.15,0.2,0.25},yticklabels={0.00,0.05,0.10,0.15,0.20,0.25},ylabel={},yticklabel style={text width=2.1em,align=right}},
colormap/viridis,
point meta max=0.252525240182877,
point meta min=0,
tick align=outside,
tick pos=left,
x grid style={white!69.0196078431373!black},
xmin=-0.5, xmax=223.5,
xtick style={color=black},
y dir=reverse,
y grid style={white!69.0196078431373!black},
ymin=-0.5, ymax=223.5,
ytick style={color=black}
]
\addplot graphics [includegraphics cmd=\pgfimage,xmin=-0.5, xmax=223.5, ymin=223.5, ymax=-0.5] {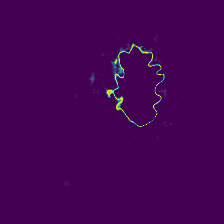};
\end{axis}

\end{tikzpicture}

%% file: alea_000246.tex
\begin{tikzpicture}

\begin{axis}[
hide axis,
axis equal image,
colorbar,
colorbar style={ytick={0,0.5,1,1.5},yticklabels={0.0,0.5,1.0,1.5},ylabel={},height=0.9*\pgfkeysvalueof{/pgfplots/parent axis width},yshift=-3mm,yticklabel style={text width=1.7em,align=right}},
colormap/viridis,
point meta max=1.5,
point meta min=0,
tick align=outside,
tick pos=left,
x grid style={white!69.0196078431373!black},
xmin=-0.5, xmax=223.5,
xtick style={color=black},
y dir=reverse,
y grid style={white!69.0196078431373!black},
ymin=-0.5, ymax=223.5,
ytick style={color=black}
]
\addplot graphics [includegraphics cmd=\pgfimage,xmin=-0.5, xmax=223.5, ymin=223.5, ymax=-0.5] {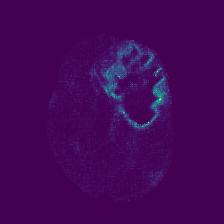};
\end{axis}

\end{tikzpicture}

%% file: bayes016867.tex
\begin{tikzpicture}

\begin{axis}[
hide axis,
axis equal image,
colorbar,
colorbar style={height=0.9*\pgfkeysvalueof{/pgfplots/parent axis width},yshift=-3mm, ytick={0,0.05,0.1,0.15,0.2,0.25},yticklabels={0.00,0.05,0.10,0.15,0.20,0.25},ylabel={},yticklabel style={text width=2.1em,align=right}},
colormap/viridis,
point meta max=0.252525240182877,
point meta min=0,
tick align=outside,
tick pos=left,
x grid style={white!69.0196078431373!black},
xmin=-0.5, xmax=223.5,
xtick style={color=black},
y dir=reverse,
y grid style={white!69.0196078431373!black},
ymin=-0.5, ymax=223.5,
ytick style={color=black}
]
\addplot graphics [includegraphics cmd=\pgfimage,xmin=-0.5, xmax=223.5, ymin=223.5, ymax=-0.5] {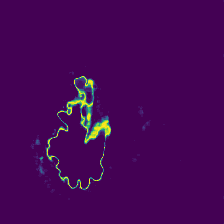};
\end{axis}

\end{tikzpicture}

%% file: alea_016867.tex
\begin{tikzpicture}

\begin{axis}[
hide axis,
axis equal image,
colorbar,
colorbar style={ytick={0,0.5,1,1.5},yticklabels={0.0,0.5,1.0,1.5},ylabel={},height=0.9*\pgfkeysvalueof{/pgfplots/parent axis width},yshift=-3mm,yticklabel style={text width=1.7em,align=right}},
colormap/viridis,
point meta max=1.5,
point meta min=0,
tick align=outside,
tick pos=left,
x grid style={white!69.0196078431373!black},
xmin=-0.5, xmax=223.5,
xtick style={color=black},
y dir=reverse,
y grid style={white!69.0196078431373!black},
ymin=-0.5, ymax=223.5,
ytick style={color=black}
]
\addplot graphics [includegraphics cmd=\pgfimage,xmin=-0.5, xmax=223.5, ymin=223.5, ymax=-0.5] {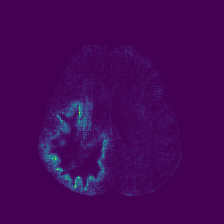};
\end{axis}

\end{tikzpicture}

%% file: bayes002468.tex
\begin{tikzpicture}

\begin{axis}[
hide axis,
axis equal image,
colorbar,
colorbar style={height=0.9*\pgfkeysvalueof{/pgfplots/parent axis width},yshift=-3mm, ytick={0,0.05,0.1,0.15,0.2,0.25},yticklabels={0.00,0.05,0.10,0.15,0.20,0.25},ylabel={},yticklabel style={text width=2.1em,align=right}},
colormap/viridis,
point meta max=0.252525240182877,
point meta min=0,
tick align=outside,
tick pos=left,
x grid style={white!69.0196078431373!black},
xmin=-0.5, xmax=223.5,
xtick style={color=black},
y dir=reverse,
y grid style={white!69.0196078431373!black},
ymin=-0.5, ymax=223.5,
ytick style={color=black}
]
\addplot graphics [includegraphics cmd=\pgfimage,xmin=-0.5, xmax=223.5, ymin=223.5, ymax=-0.5] {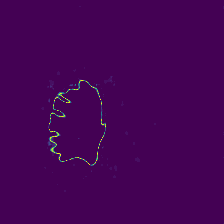};
\end{axis}

\end{tikzpicture}

%% file: alea_002468.tex
\begin{tikzpicture}

\begin{axis}[
hide axis,
axis equal image,
colorbar,
colorbar style={ytick={0,0.5,1,1.5},yticklabels={0.0,0.5,1.0,1.5},ylabel={},height=0.9*\pgfkeysvalueof{/pgfplots/parent axis width},yshift=-3mm,yticklabel style={text width=1.7em,align=right}},
colormap/viridis,
point meta max=1.5,
point meta min=0,
tick align=outside,
tick pos=left,
x grid style={white!69.0196078431373!black},
xmin=-0.5, xmax=223.5,
xtick style={color=black},
y dir=reverse,
y grid style={white!69.0196078431373!black},
ymin=-0.5, ymax=223.5,
ytick style={color=black}
]
\addplot graphics [includegraphics cmd=\pgfimage,xmin=-0.5, xmax=223.5, ymin=223.5, ymax=-0.5] {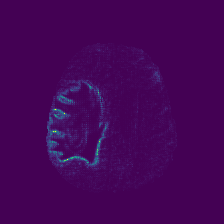};
\end{axis}

\end{tikzpicture}

%% file: mean013856.tex
\begin{tikzpicture}

\begin{axis}[
hide axis,
axis equal image,
colorbar,
colorbar style={height=0.9*\pgfkeysvalueof{/pgfplots/parent axis width},ytick={0,0.2,0.4,0.6,0.8,1},yticklabels={0.0,0.2,0.4,0.6,0.8,1.0},ylabel={}, yshift=-3mm,yticklabel style={text width=1.7em,align=right}},
colormap/viridis,
point meta max=1,
point meta min=0,
tick align=outside,
tick pos=left,
x grid style={white!69.0196078431373!black},
xmin=-0.5, xmax=223.5,
xtick style={color=black},
y dir=reverse,
y grid style={white!69.0196078431373!black},
ymin=-0.5, ymax=223.5,
ytick style={color=black}
]
\addplot graphics [includegraphics cmd=\pgfimage,xmin=-0.5, xmax=223.5, ymin=223.5, ymax=-0.5] {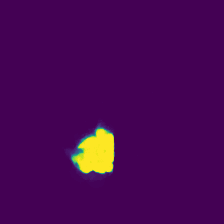};
\end{axis}

\end{tikzpicture}

%% file: var013856.tex
\begin{tikzpicture}

\begin{axis}[
hide axis,
axis equal image,
colorbar,
colorbar style={ytick={0,0.05,0.1,0.15,0.2,0.25},yticklabels={0.00,0.05,0.10,0.15,0.20,0.25},ylabel={},height=0.9*\pgfkeysvalueof{/pgfplots/parent axis width},yshift=-3mm,yticklabel style={text width=2.1em,align=right}},
colormap/viridis,
point meta max=0.252525240182877,
point meta min=0,
tick align=outside,
tick pos=left,
x grid style={white!69.0196078431373!black},
xmin=-0.5, xmax=223.5,
xtick style={color=black},
y dir=reverse,
y grid style={white!69.0196078431373!black},
ymin=-0.5, ymax=223.5,
ytick style={color=black}
]
\addplot graphics [includegraphics cmd=\pgfimage,xmin=-0.5, xmax=223.5, ymin=223.5, ymax=-0.5] {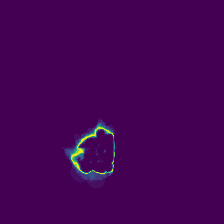};
\end{axis}

\end{tikzpicture}

%% file: mean001919.tex
\begin{tikzpicture}

\begin{axis}[
hide axis,
axis equal image,
colorbar,
colorbar style={height=0.9*\pgfkeysvalueof{/pgfplots/parent axis width},ytick={0,0.2,0.4,0.6,0.8,1},yticklabels={0.0,0.2,0.4,0.6,0.8,1.0},ylabel={}, yshift=-3mm,yticklabel style={text width=1.7em,align=right}},
colormap/viridis,
point meta max=1,
point meta min=0,
tick align=outside,
tick pos=left,
x grid style={white!69.0196078431373!black},
xmin=-0.5, xmax=223.5,
xtick style={color=black},
y dir=reverse,
y grid style={white!69.0196078431373!black},
ymin=-0.5, ymax=223.5,
ytick style={color=black}
]
\addplot graphics [includegraphics cmd=\pgfimage,xmin=-0.5, xmax=223.5, ymin=223.5, ymax=-0.5] {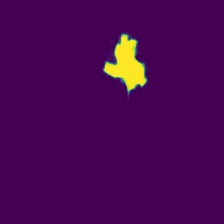};
\end{axis}

\end{tikzpicture}

%% file: var001919.tex
\begin{tikzpicture}

\begin{axis}[
hide axis,
axis equal image,
colorbar,
colorbar style={ytick={0,0.05,0.1,0.15,0.2,0.25},yticklabels={0.00,0.05,0.10,0.15,0.20,0.25},ylabel={},height=0.9*\pgfkeysvalueof{/pgfplots/parent axis width},yshift=-3mm,yticklabel style={text width=2.1em,align=right}},
colormap/viridis,
point meta max=0.252525240182877,
point meta min=0,
tick align=outside,
tick pos=left,
x grid style={white!69.0196078431373!black},
xmin=-0.5, xmax=223.5,
xtick style={color=black},
y dir=reverse,
y grid style={white!69.0196078431373!black},
ymin=-0.5, ymax=223.5,
ytick style={color=black}
]
\addplot graphics [includegraphics cmd=\pgfimage,xmin=-0.5, xmax=223.5, ymin=223.5, ymax=-0.5] {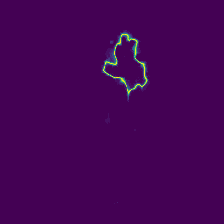};
\end{axis}

\end{tikzpicture}

%% file: mean004368.tex
\begin{tikzpicture}

\begin{axis}[
hide axis,
axis equal image,
colorbar,
colorbar style={height=0.9*\pgfkeysvalueof{/pgfplots/parent axis width},ytick={0,0.2,0.4,0.6,0.8,1},yticklabels={0.0,0.2,0.4,0.6,0.8,1.0},ylabel={}, yshift=-3mm,yticklabel style={text width=1.7em,align=right}},
colormap/viridis,
point meta max=1,
point meta min=0,
tick align=outside,
tick pos=left,
x grid style={white!69.0196078431373!black},
xmin=-0.5, xmax=223.5,
xtick style={color=black},
y dir=reverse,
y grid style={white!69.0196078431373!black},
ymin=-0.5, ymax=223.5,
ytick style={color=black}
]
\addplot graphics [includegraphics cmd=\pgfimage,xmin=-0.5, xmax=223.5, ymin=223.5, ymax=-0.5] {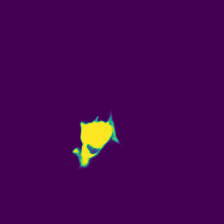};
\end{axis}

\end{tikzpicture}

%% file: var004368.tex
\begin{tikzpicture}

\begin{axis}[
hide axis,
axis equal image,
colorbar,
colorbar style={ytick={0,0.05,0.1,0.15,0.2,0.25},yticklabels={0.00,0.05,0.10,0.15,0.20,0.25},ylabel={},height=0.9*\pgfkeysvalueof{/pgfplots/parent axis width},yshift=-3mm,yticklabel style={text width=2.1em,align=right}},
colormap/viridis,
point meta max=0.252525240182877,
point meta min=0,
tick align=outside,
tick pos=left,
x grid style={white!69.0196078431373!black},
xmin=-0.5, xmax=223.5,
xtick style={color=black},
y dir=reverse,
y grid style={white!69.0196078431373!black},
ymin=-0.5, ymax=223.5,
ytick style={color=black}
]
\addplot graphics [includegraphics cmd=\pgfimage,xmin=-0.5, xmax=223.5, ymin=223.5, ymax=-0.5] {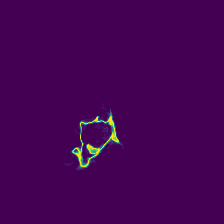};
\end{axis}

\end{tikzpicture}